\documentclass{article}
\usepackage{spconf,amsmath,graphicx}
\usepackage{tipa}

\title{HIERARCHICAL NEURAL NETWORK ARCHITECTURE IN KEYWORD SPOTTING}
\name{Yixiao Qu, Sihao Xue, Zhenyi Ying, Hang Zhou, Jue Sun}
\address{NIO Co., Ltd\\
	      \emph{\{yixiao.qu, sihao.xue, zhenyi.ying, hang.zhou, jue.sun\}@nio.com}}
\begin{document}
%\ninept
%
\maketitle
\begin{abstract}
Keyword Spotting (KWS) provides the start signal of ASR problem, and thus it is essential to ensure a high recall rate. However, its real-time property requires low computation complexity. This contradiction inspires people to find a suitable model which is small enough to perform well in multi environments.

To deal with this contradiction, we implement the Hierarchical Neural Network(HNN), which is proved to be effective in many speech recognition problems. HNN outperforms traditional DNN and CNN even though its model size and computation complexity are slightly less. Also, its simple topology structure makes easy to deploy on any device.

\end{abstract}
\begin{keywords}
Hierarchical Neural Network, Keyword Spotting, Bottleneck Feature
\end{keywords}
\section{Introduction}
\label{sec:intro}

In recent years, significant progress has been achieved in the speech recognition field, and researchers focus more on real-time problems such as keyword spotting (KWS). Most of the time, KWS provides the start signal of ASR problem. Thus, it requires a high recall rate, low mistake rate, and fast calculation speed. Although large and complex model guarantees high recall and low mistake rate, it also requires substantial computing power. Therefore, many complex models like LSTM are not suitable for KWS problem.

Traditionally, deep neural network (DNN) and convolutional neural network (CNN) are adopted to solve KWS problem and have achieved excellent performance. However, some KWS problems require the model to perform well in various scenarios under different background noise, and many simple KWS models fail to meet this need.

Motivated by this problem, we implemented the Hierarchical Neural Network (HNN) in KWS problem. HNN uses several models trained at different levels to calculate posterior probability together. The final result depends on the combination of all the networks from different levels. Each model is trained with different scenario training data. Low-level models provide bottleneck features to the high-level models. In our experiment, this architecture shows considerable improvement over the traditional network.

Section 2 describes some related work. Section 3 talks about KWS algorithm used by us. Sections 4 introduces HNN and experiments are described in section 5. The paper concludes with a discussion in section 6.
%These guidelines include complete descriptions of the fonts, spacing, and
%related information for producing your proceedings manuscripts. Please follow
%them and if you have any questions, direct them to Conference Management
%Services, Inc.: Phone +1-979-846-6800 or email
%to \\\texttt{icassp2019@cmsworkshops.com}.

\section{Related Work}
\label{sec:format}

 Thomas et al. [1] uses multilingual MLP features to build a Large Vocabulary Continuous Speech Recognition (LVSCR) systems. Plahl et al. [2][3] applies hierarchical bottleneck feature for LVSCR. Valente et al. [4] realizes hierarchical processing spectrum for mandarin LVCSR system. 

There are also plenty of literature on the topic of KWS. Offline LVCSR systems can be used for detecting the keywords of interest. [5][6]. Moreover, Hidden Markov Models (HMM) are commonly used for online KWS system [7][8]. In traditional, Gaussian Mixture Models (GMM) is used in acoustic modeling under the HMM framework. It is replaced by Deep Neural Network (DNN) with time goes on [9]. And several architectures have been applied [10][11].

\section{Keyword spotting}
\label{sec:format}

In general, KWS is implemented on local devices and is a real-time problem, therefore low latency and memory storage are required to ensure user experience and acceptable consumption. The early KWS is based on offline continuous speech recognition with GMM-HMM [7][8]. With the great success of Deep Neural Network in continuous speech recognition, traditional GMM-HMM is replaced by DNN [9]. Recently, Chen er al. [12] designed a KWS strategy without HMM.

In our research, we use finite state transducer (FST) to realize KWS by employing work unit. FST consists of a finite number of states. Each state is connected by a transition labeled with input/output pair. State transitions depend on the inputs and transition rules. For example, we use ``hi nomi", which is implemented in our product, as a keyword. FST will begin with searching ``hi nomi" in the dictionary for its phone units. Its pronunciation is ``hai n\textschwa\textupsilon mi" and we choose ``HH AY1 N OW1 M IY1" as its phone state. Then find its all tri-phones such as ``HH-AY1-N" (which may occur in speech data) and do clustering to generate each state. During clustering, we let the tri-phones whose central phone is ``HH" (like ``HH-HH-AY1" and ``sil-HH-AY1" etc. ) and ``AY1" (like ``HH-AY1-AY1" and ``AY1-AY1-AY1" etc.) as the first word state,  ``HH-AH1"; ``N" and ``OW1" as the second word state ``N-OW1"; ``M" and ``IY1" as the third word state ``M-IY1". ``sil" is the silent state and any input which does not occur on the connected arcs is directed to the ``other" state. These labels are generated via forced alignment using our LVSCR system. The ``hi nomi" FST is shown in Figure.1. The expression is input/output pairs, such as ``HH-AY1",  and the arrow means state transformation. Device wakes up when the output equals 1. It would wake up if and only if ``hi nomi" occurs.

\begin{figure}[htb]

\begin{minipage}[b]{1.0\linewidth}
 \centering
  \centerline{\includegraphics[width=4cm]{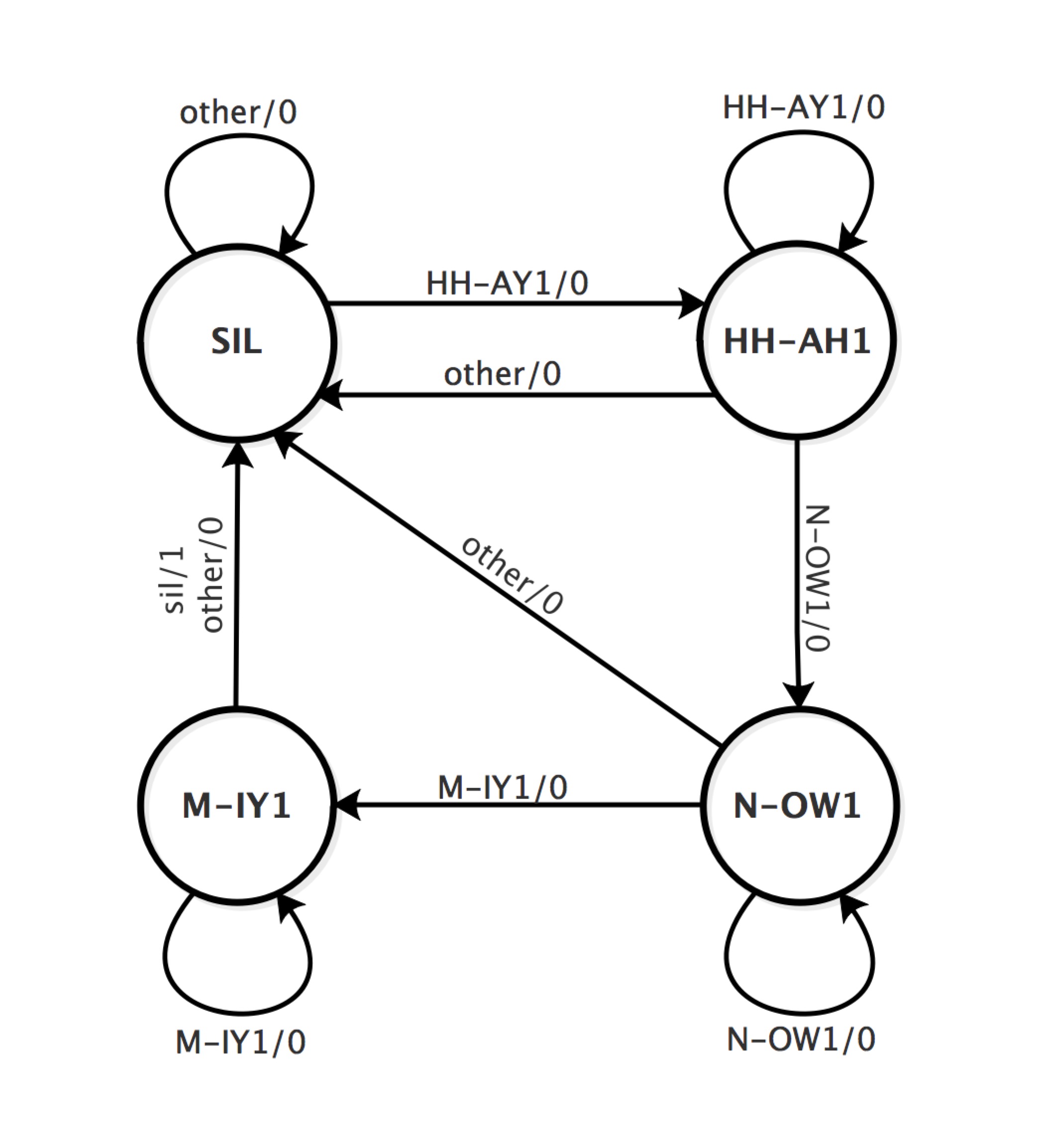}}
 % \vspace{2.0cm}
%  \centerline{(a) Result 1}\medskip
\end{minipage}

\caption{FST for keyword ``hi nomi".}%
\label{fig:res}

\end{figure}

\section{Hierarchical Neural Network}
\label{sec:pagestyle}

In this section, we elaborate how we realize the Hierarchical Neural Network(HNN) in KWS problem

\subsection{Training Neural Network}
\label{ssec:subhead}
The neural network is trained on three environment training data - quiet (Q), video (V) and incar (C). The major difference between these three datasets is background noise. Quiet data has minimal background noise; video data has noise that is extracted from videos such as movie or television programs; noise for incar data consists of road noise and external noise generated from a moving vehicle.  The network is trained in the following steps:

A. Train the MLP on quiet training data. The first level network is a traditional neural network with bottleneck BN1 and randomly initialized weight. The bottleneck layer allows the network to learn low dimensional feature extracted among quiet environments data.

B. Training the second level network on video training data. Apart from input feature dimension, it shares the same architecture as the first level. Besides the original input feature, it also inputs the bottleneck feature BN1 from the first level network. Also, the second level network has a bottleneck layer BN2.

C. Training the third level network on incar training data. The input of third level network includes original input feature and bottleneck feature. Unlike the first two level network, the third level network is the last one, and it does not have bottleneck architecture.

And the final HNN is the three-level network combined. The training architecture of HNN is shown in Figure.2.

\begin{figure}[htb]

\begin{minipage}[b]{1.0\linewidth}
 \centering
  \centerline{\includegraphics[width=6cm]{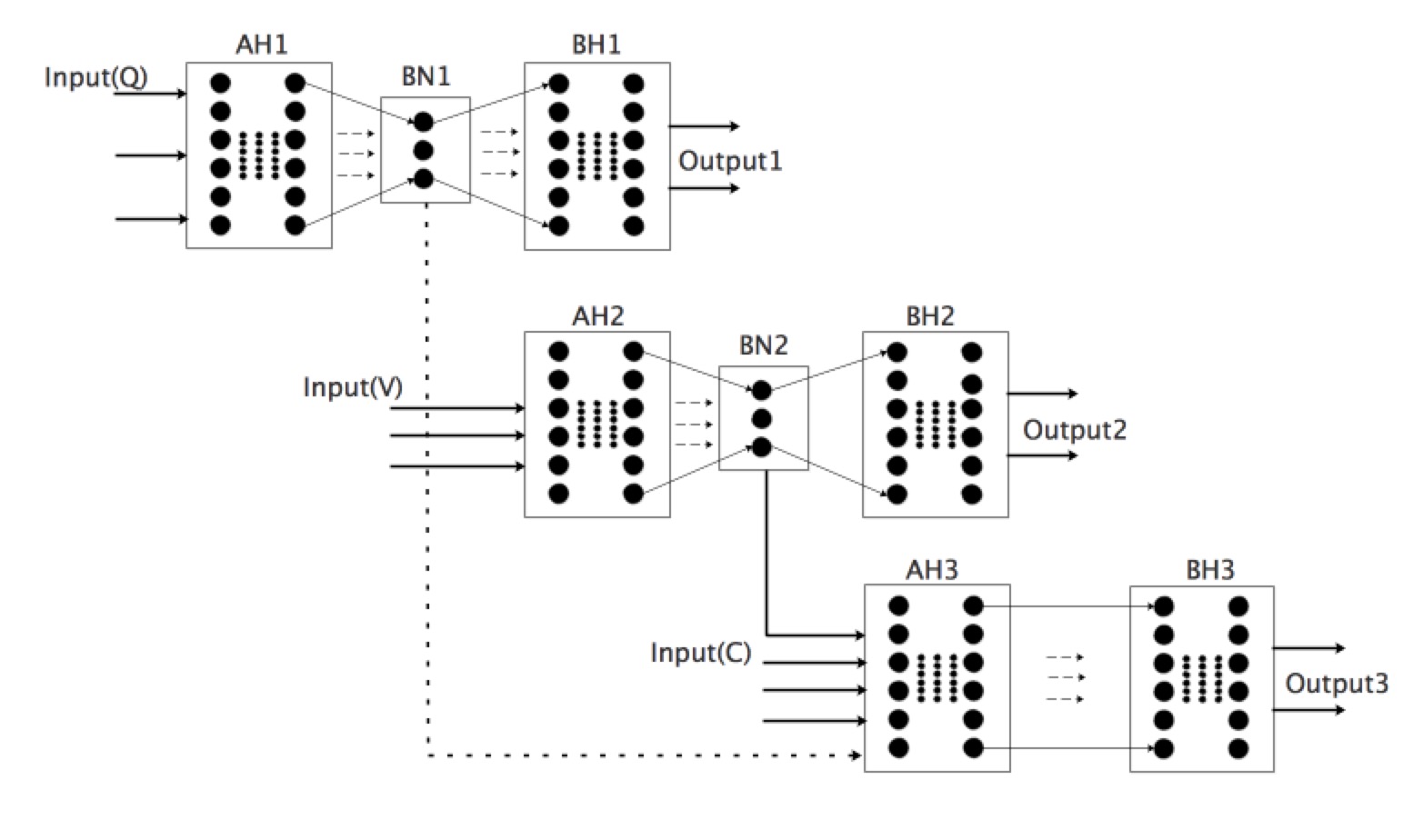}}
 % \vspace{2.0cm}
%  \centerline{(a) Result 1}\medskip
\end{minipage}

\caption{HNN Architecture. BN1 BN2 is the bottleneck of first two level. The dotted arrow means the bottleneck feature is optional for third level. And different level network is trained with different environment training data.}%
\label{fig:res}

\end{figure}

\subsection{Post Processing and Wakeup Decision}

Post processing is another problem of HNN because three level model have the same input but different outputs . 

Obviously, there are two post processing ways, only retaining the third level output and retaining all outputs. The configuration is showed in Figure.3. 

\begin{figure}[htb]

\begin{minipage}[b]{1.0\linewidth}
 \centering
  \centerline{\includegraphics[width=6cm]{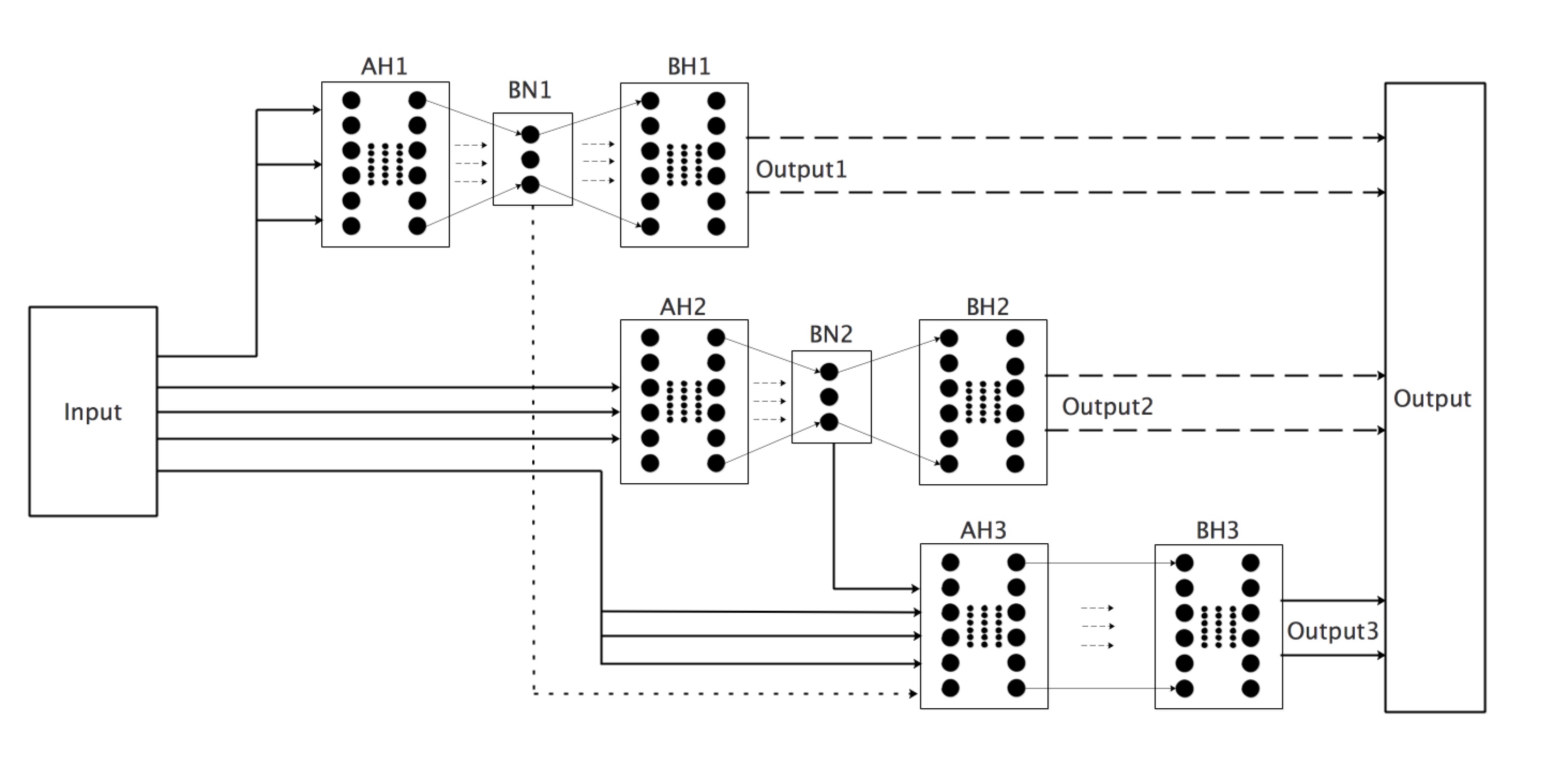}}
 % \vspace{2.0cm}
%  \centerline{(a) Result 1}\medskip
\end{minipage}

\caption{Two output processing method. The dotted arrow of output means the output is optional when the final output is calculated. And all level network use same input feature}%
\label{fig:res}

\end{figure}

There are also two strategies when retaining all level outputs. 

A. Wake up the device as long as any level neural network wakes up the device.

B. Use the average of three level model outputs to decide whether the device wake up.

\label{ssec:subhead}

\section{Experiment}
\label{sec:typestyle}

We implement our HNN algorithm in KWS problem. We use ``hi nomi" as the wake-up keyword. Video and incar noise is mixed into the training data. The training data consists of  830k utterances which include 520k positive samples and 310k negative samples.180 hours of data is used to test the recall rate and false alarm (FA). Besides 20k keyword utterances, the test data includes various environmental data such as conference, incar, video and etc. We use KALDI toolkit [13] to train each model in our experiment.

The baseline model is a 3-layer DNN with 4 outputs. Each hidden layer has 512 nodes. We generate acoustic feature based on 40-dimensional log-filterbank energies computed every 10ms over a window of 25ms. The input context is set to 11 so that the format is ``5-1-5". For comparison, the HNN in our experiment has similar model size and calculation complexity with the baseline model. The model architecture and calculation complexity are showed in Table.1 and Table .2.

\begin{table}[th]

  \centering
  \begin{tabular}{ c| c| c| c|}
    \hline
 %   \toprule
    \multicolumn{1}{c|}{\textbf{}} & 
                                         \multicolumn{1}{c|}{\textbf{1st lv}} &
                                         \multicolumn{1}{c|}{\textbf{2nd lv}} &
                                         \multicolumn{1}{c|}{\textbf{3rd lv}}\\
  \hline
%    \midrule
    $HNN1AH$                  & ~~~$2 \times 256$~~~    & $2 \times 256$    & $3\times 256$     \\
    $HNN1BN$                  & ~~$128$~~~      & ~~$128$~~~  & ~~$-$~~~         \\
    $HNN1BH$                  & ~~~$2 \times 256$~~~    & ~~~$2 \times 256$~~~ & ~~$-$~~~         \\    
   \hline
    $HNN2AH$                  & ~~~$2 \times 256$~~~    & $2 \times 256$    & $3\times 256$     \\
    $HNN2BN$                  & ~~$64$~~~      & ~~$64$~~~  & ~~$-$~~~         \\
    $HNN2BH$                  & ~~~$2 \times 256$~~~    & ~~~$2\times 256$~~~ & ~~$-$~~~         \\  
  \hline
    $HNN3AH$                  & ~~~$2 \times 256$~~~    & $1 \times 256$    & $2\times 256$     \\
    $HNN3BN$                  & ~~$128$~~~      & ~~$128$~~~  & ~~$-$~~~         \\
    $HNN3BH$                  & ~~~$2 \times 256$~~~    & ~~~$1 \times 256$~~~ & ~~$-$~~~         \\  
  \hline
  %  \bottomrule
  \end{tabular}
    \caption{HNN Architecture. AH is the prior hidden layer of bottleneck layer, BH is the posterior hidden layer of bottleneck layer. }
  \label{tab:example}
\end{table}

\begin{table}[th]

  \centering
  \begin{tabular}{ c| c| c| c|}
  \hline
 %   \toprule
    \multicolumn{1}{c|}{\textbf{}} & 
                                         \multicolumn{1}{c|}{\textbf{all bn}} &
                                         \multicolumn{1}{c|}{\textbf{all output}} &
                                         \multicolumn{1}{c|}{\textbf{calculation}}\\
%    \midrule
  \hline
    $BASELINE$                  & $-$   & $-$   &~~~ $\sim 0.97M $   \\
  \hline
    $HNN1$                  & $YES$   & $YES$   &~~~ $\sim 1.28M $   \\
    $HNN1$                  & ~$NO$~~~      & ~~~~$YES$~~~  & ~~~~~~$\sim 1.26M$~~~         \\
    $HNN1$                  & $YES$   & $NO$   &~~ $\sim 1.09M $   \\
    $HNN1$                  & ~$NO$~~~      & ~~~$NO$~~~  & ~~~~~~$\sim 1.05M$~~~         \\
    \hline
    $HNN2$                  & $YES$   & $YES$   &~~ $\sim 1.19M $   \\
    $HNN2$                  & ~$NO$~~~      & ~~~~$YES$~~~  & ~~~~~~$\sim 1.15M$~~~         \\
    $HNN2$                  & $YES$   & $NO$   &~~ $\sim 1.00M $   \\
    $HNN2$                  & ~$NO$~~~      & ~~~$NO$~~~  & ~~~~~$\sim 0.99M$~~~         \\
      \hline
    $HNN3$                  & $YES$   & $YES$   &~ $\sim 1.09M $   \\
    $HNN3$                  & ~$NO$~~~      & ~~~~$YES$~~~  & ~~~~~$\sim 1.06M$~~~         \\
    $HNN3$                  & $YES$   & $NO$   &~ $\sim 0.98M $   \\
    $HNN3$                  & ~$NO$~~~      & ~~~$NO$~~~  & ~~~~~$\sim 0.95M$~~~         \\
  \hline
  %  \bottomrule
  \end{tabular}
    \caption{HNN Computation Complexity. All bn means the third get all first two level bottleneck feature as part of inputs. All output means the network calculate final output by combining all three level output}
  \label{tab:example}
\end{table}

\subsection{Bottleneck Architecture}
\label{ssec:subhead}

The three levels of the HNN are trained with quiet(Q), video(V) and incar(C) keyword training data respectively. We compare the performance of HNN with different bottleneck architecture.

The ROC curve is shown in Figure.4 and Figure.5. According to the ROC, HNN performs much better than the baseline DNN model. What's more, no matter how we process the three-level outputs, 1 bottleneck architecture performs better than all bottleneck architecture. The reason may be that the first level bottleneck feature is included in the second level bottleneck feature. So more bottleneck feature input, more parameter has to be trained. Moreover, these parameters may influence convergence.

\begin{figure}[htb]

\begin{minipage}[b]{1.0\linewidth}
 \centering
  \centerline{\includegraphics[width=5cm]{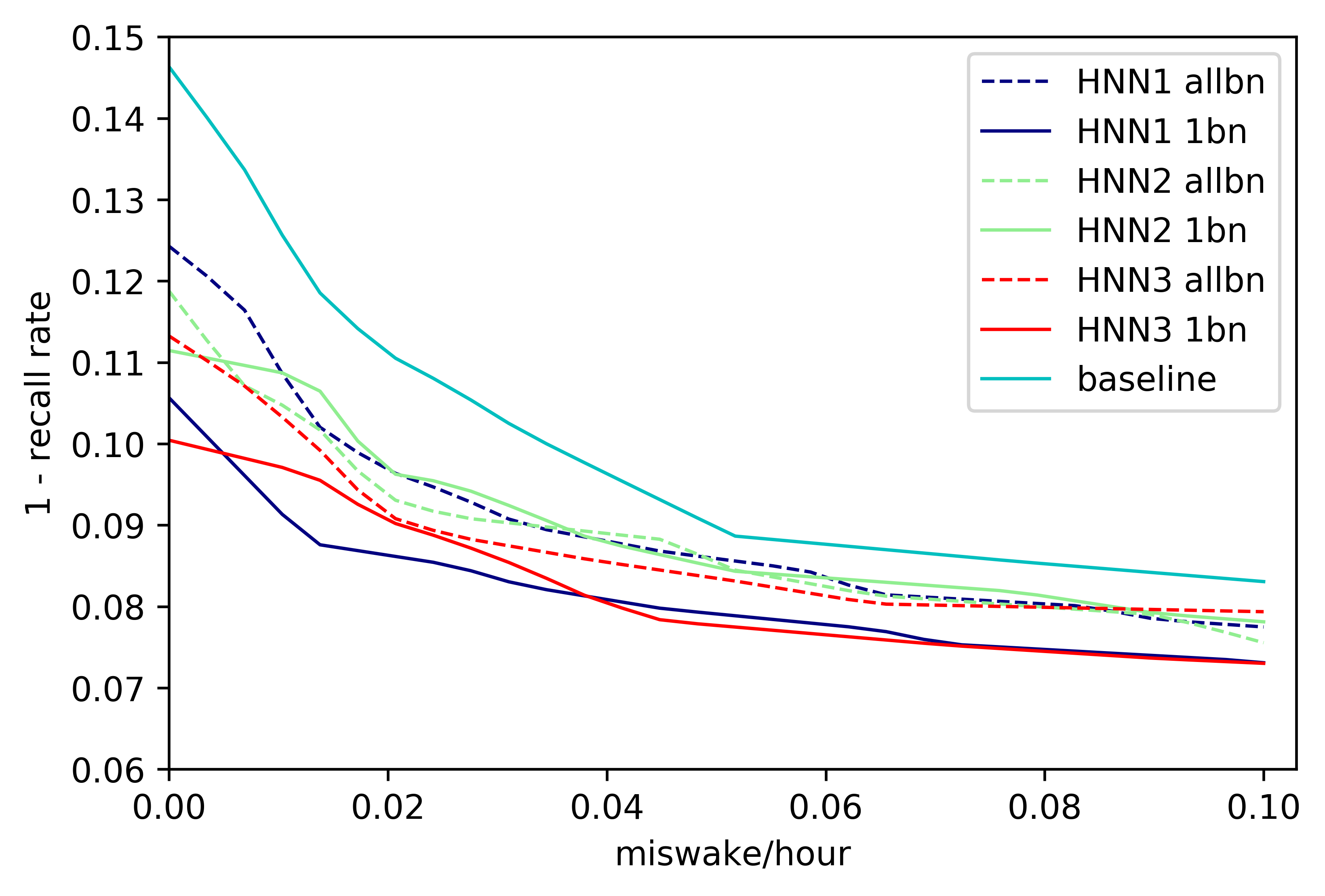}}
 % \vspace{2.0cm}
%  \centerline{(a) Result 1}\medskip
\end{minipage}

\caption{The ROC curve of HNN. HNN only keeps the third level output}%
\label{fig:res}

\end{figure}

\begin{figure}[htb]

\begin{minipage}[b]{1.0\linewidth}
 \centering
  \centerline{\includegraphics[width=5cm]{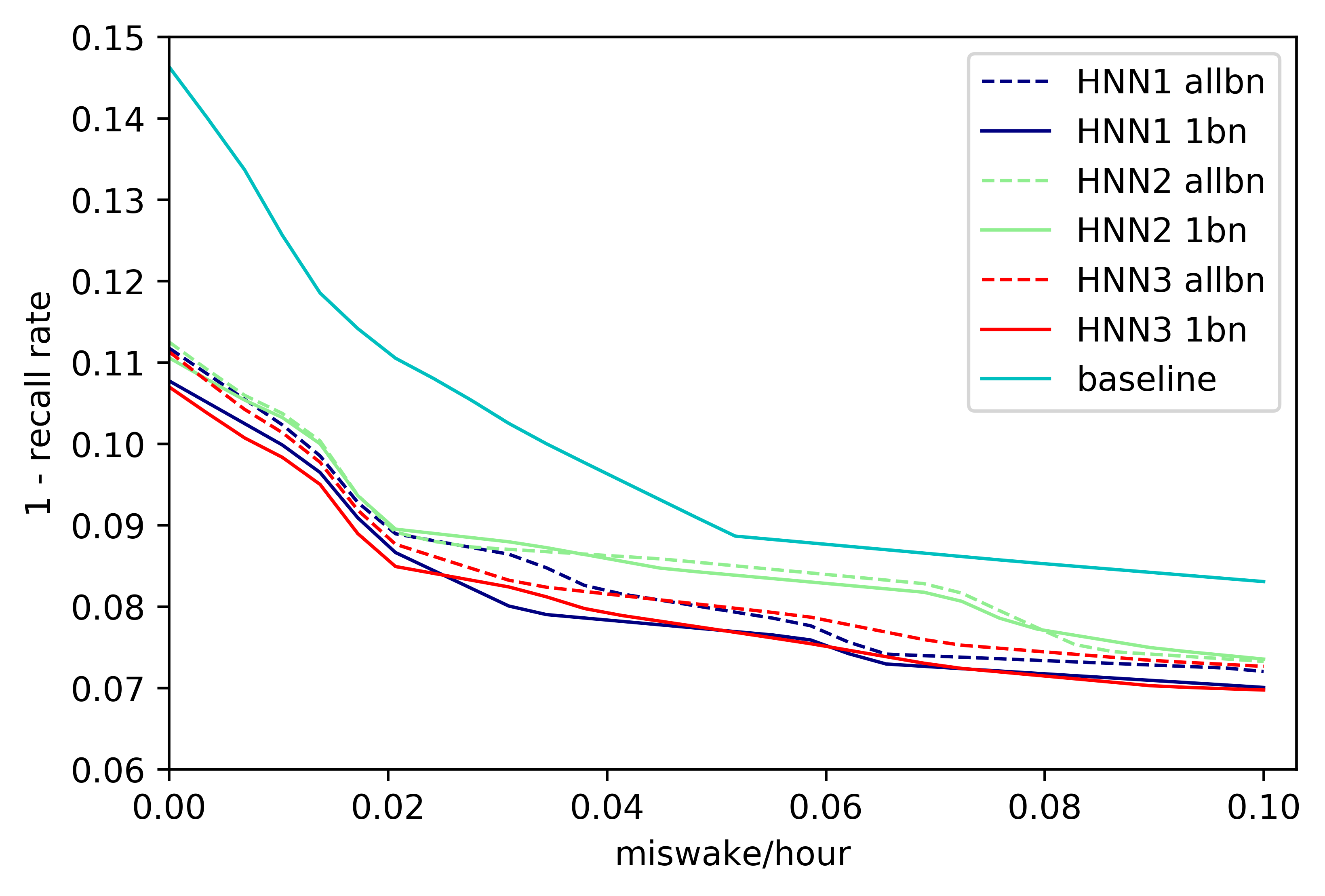}}
 % \vspace{2.0cm}
%  \centerline{(a) Result 1}\medskip
\end{minipage}

\caption{The ROC curve of HNN. It wakes up as long as any level network wakes up}%
\label{fig:res}

\end{figure}

%\begin{figure}[htb]
%\begin{minipage}[b]{.48\linewidth}
 % \centering
 % \centerline{\includegraphics[width=4.0cm]{1output.jpg}}
 % \vspace{1.5cm}
% \centerline{(1)  }\medskip
%\end{minipage}
%!TEX encoding = UTF-8 Unicode%\hfill
%\begin{minipage}[b]{0.48\linewidth}
% \centering
%  \centerline{\includegraphics[width=4.0cm]{alloutput.jpg}}
 % \vspace{1.5cm}
 % \centerline{(2)  }\medskip
%\end{minipage}

%\caption{ROC curve of Hierarchical Neural Network. Hierarchical Neural Networks in (1) only keep the third level output. Hierarchical Neural Network in (2) wakeup the device as long as any level network wakes up the device}%
%\label{fig:res}

%\end{figure}

\subsection{Output Process}
\label{ssec:subhead}

Besides bottleneck architecture, there are also three ways to process output:

A. only keep the final level output

B. wakeup the device as long as any network at any level wakes up the device

C. calculate the average of three-level network's posterior to decide whether or not wake up the device.

We find that the performance of B and C is similar. So we only show the result of A and B in this paper. The ROC is shown in Figure.6. In this part experiment, we only use 1 bottleneck HNN as the result in 5.1.

\begin{figure}[htb]

\begin{minipage}[b]{1.0\linewidth}
 \centering
  \centerline{\includegraphics[width=5cm]{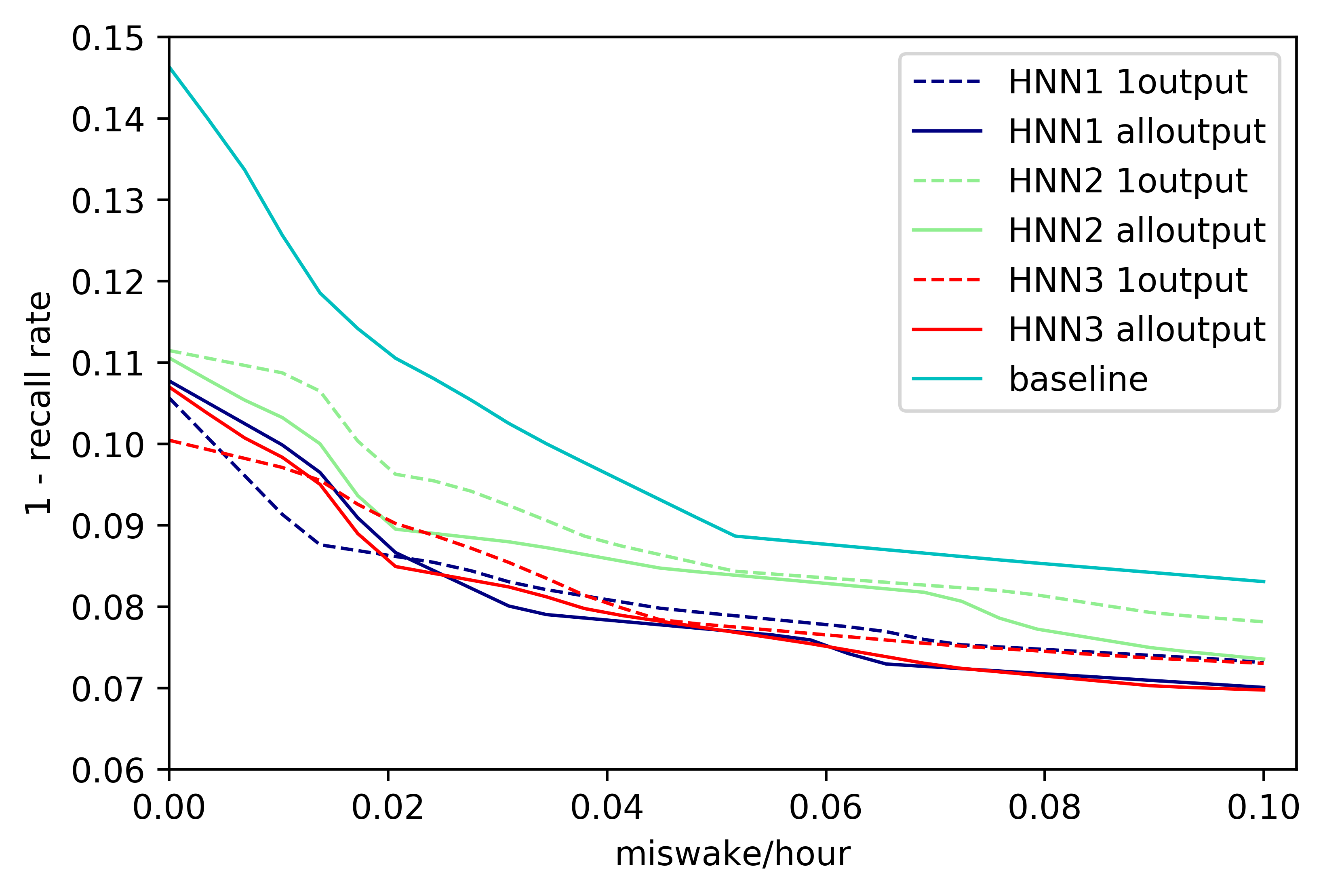}}
 % \vspace{2.0cm}
%  \centerline{(a) Result 1}\medskip
\end{minipage}

\caption{ROC curve of HNN. All network use 1 bottleneck architecture.}%
\label{fig:res}

\end{figure}

\begin{figure}[htb]

\begin{minipage}[b]{1.0\linewidth}
 \centering
  \centerline{\includegraphics[width=5cm]{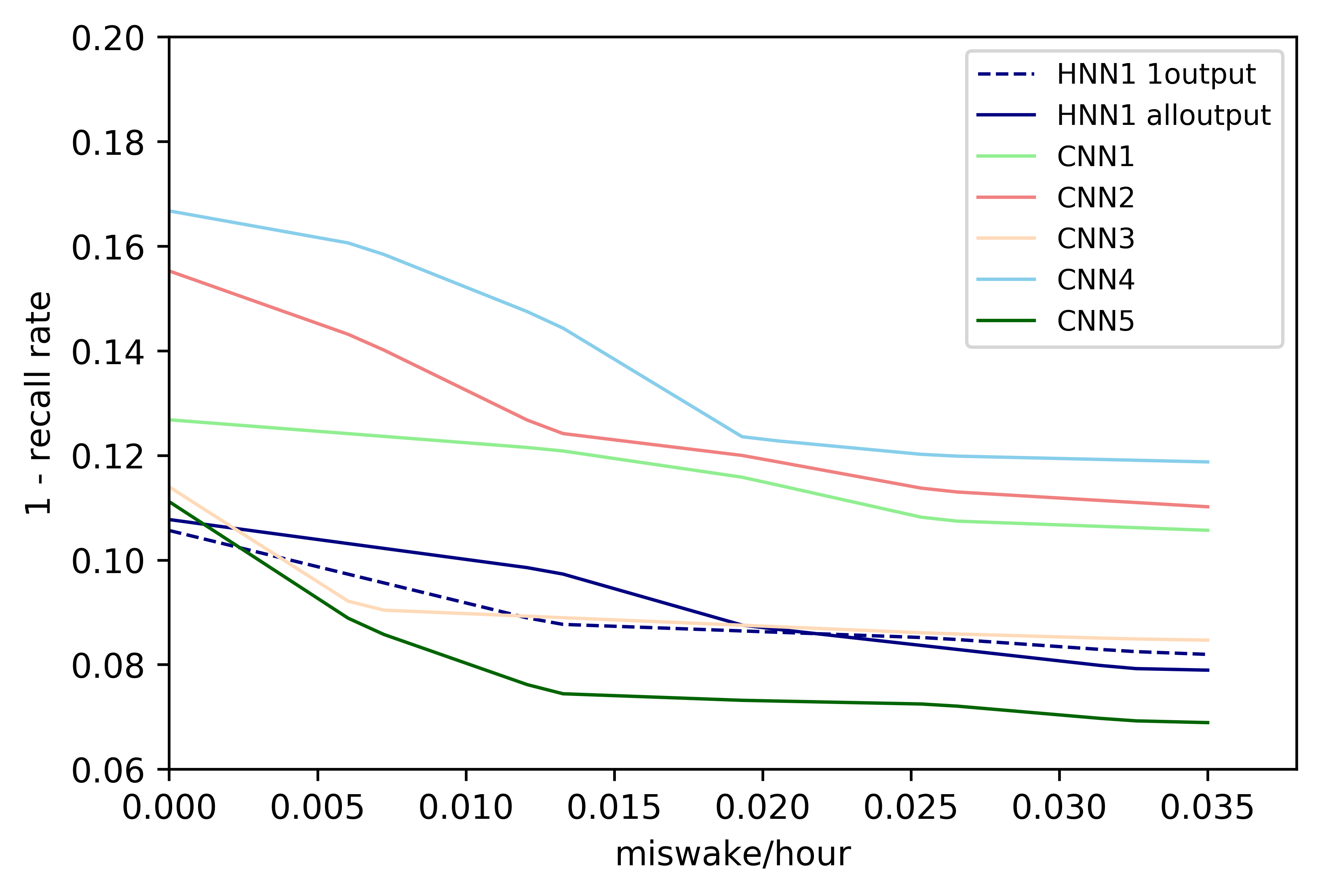}}
 % \vspace{2.0cm}
%  \centerline{(a) Result 1}\medskip
\end{minipage}

\caption{ROC curve of HNN. All network use 1 bottleneck architecture.}%
\label{fig:res}

\end{figure}

At most of the time, the performance of all output HNN is better than 1 output architecture, but it performs worse when FA is low. It is easy to imagine that using three-level outputs can increase the recall rate. When FA is low, all level networks are required to have good performance. If any level of the HNN makes a mistake, the network will generate a false alarm.

\subsection{Compare with CNN}
\label{ssec:subhead}

Besides comparing with DNN baseline, we also examine the performance of HNN with some CNN baseline. HNN1 is chosen to compare with CNNs because it has the best performance according to the previous experiments. The CNN architecture is shown in Table.3. The ROC is shown in Figure.7.

According to ROC, HNN performs better than most CNN we get. But some CNN is still better than HNN.

\begin{table}[th]

  \centering
  \begin{tabular}{ c| c| c| c|}
  \hline
 %   \toprule
    \multicolumn{1}{c|}{\textbf{}} & 
                                         \multicolumn{1}{c|}{\textbf{Kernel}} &
                                         \multicolumn{1}{c|}{\textbf{Affine}} &
                                         \multicolumn{1}{c|}{\textbf{calculation}}\\
  \hline
%    \midrule
    $CNN1$                  & $64 \times [5, 7], S[2, 2]$   & $3 \times 512$   &~ $\sim 1.03M $   \\
    $CNN2$                  & $32 \times [5, 5], S[2, 3]$   & $3 \times 512$   &~~$\sim 0.87M $   \\
    $CNN3$                  & $64 \times [5, 5], S[2, 2]$   & $3 \times 512$  & ~~$\sim 1.03M$        \\
    $CNN4$                  & $32 \times [5, 5], S[2, 1]$   & $3 \times 512$   &~ $\sim 1.03M $   \\
    $CNN5$                  & $64 \times [5, 5], S[2, 3]$   & $3 \times 512$  & ~~$\sim 0.95M$   \\
  
  \hline

  %  \bottomrule
  \end{tabular}
    \caption{CNN baseline architecture, the format in Kernel is  Number of Kernel $\times$ [Kernel Length, Kernel Height], S[Length Stride, Height Stride]}
  \label{tab:example}
\end{table}

\subsection{Multi Hierarchical Neural Network}
\label{ssec:subhead}

According to CNN performance and calculation complexity, we implement Multi Hierarchical Neural Network(MHNN) whose first two level is CNN and third level is DNN. We chose same kernel as CNN5 for CNN layer. And the affine configuration of MHNN and computation complexity is showed in Table.4 and Table.5. And the ROC is showed in Figure.8.

\begin{table}[th]

  \centering
  \begin{tabular}{ c| c| c| c|}
    \hline
 %   \toprule
    \multicolumn{1}{c|}{\textbf{}} & 
                                         \multicolumn{1}{c|}{\textbf{1st lv}} &
                                         \multicolumn{1}{c|}{\textbf{2nd lv}} &
                                         \multicolumn{1}{c|}{\textbf{3rd lv}}\\
  \hline
%    \midrule

    $MHNNAH$                  & ~~~$512$~~~    & $ 512$    & $3 \times 256$     \\
    $HNN1BN$                  & ~~~$128$~~~      & ~~~$128$~~~  & ~~$-$~~~         \\
    $HNN1BH$                  & ~~~$512$~~~    & ~~~$512$~~~ & ~~$-$~~~         \\    
   \hline

  %  \bottomrule
  \end{tabular}
    \caption{MHNN architecture. The parameter of MHNNAH of level 1 and 2 is dimension of maxpooling}
  \label{tab:example}
\end{table}

\begin{table}[th]

  \centering
  \begin{tabular}{ c| c| c| c|}
  \hline
 %   \toprule
    \multicolumn{1}{c|}{\textbf{}} & 
                                         \multicolumn{1}{c|}{\textbf{all bn}} &
                                         \multicolumn{1}{c|}{\textbf{all output}} &
                                         \multicolumn{1}{c|}{\textbf{calculation}}\\
%    \midrule
  \hline
    $CNN5$                  & $-$   & $-$   &~~ $\sim 0.95M $   \\
  \hline
    $MCNN1$                  & $YES$   & $YES$   &~~ $\sim 1.15M $   \\
    $MCNN1$                  & ~$NO$~~~      & ~~~~$YES$~~~  & ~~~~~~$\sim 1.12M$~~~         \\
    $MCNN1$                  & $YES$   & $NO$   &~~ $\sim 1.00M $   \\
    $MCNN1$                  & ~$NO$~~~      & ~~~$NO$~~~  & ~~$\sim 0.97$~~~         \\

  \hline
  %  \bottomrule
  \end{tabular}
    \caption{MHNN Computation Complexity.}
  \label{tab:example}
\end{table}

\begin{figure}[htb]

\begin{minipage}[b]{1.0\linewidth}
 \centering
  \centerline{\includegraphics[width=5cm]{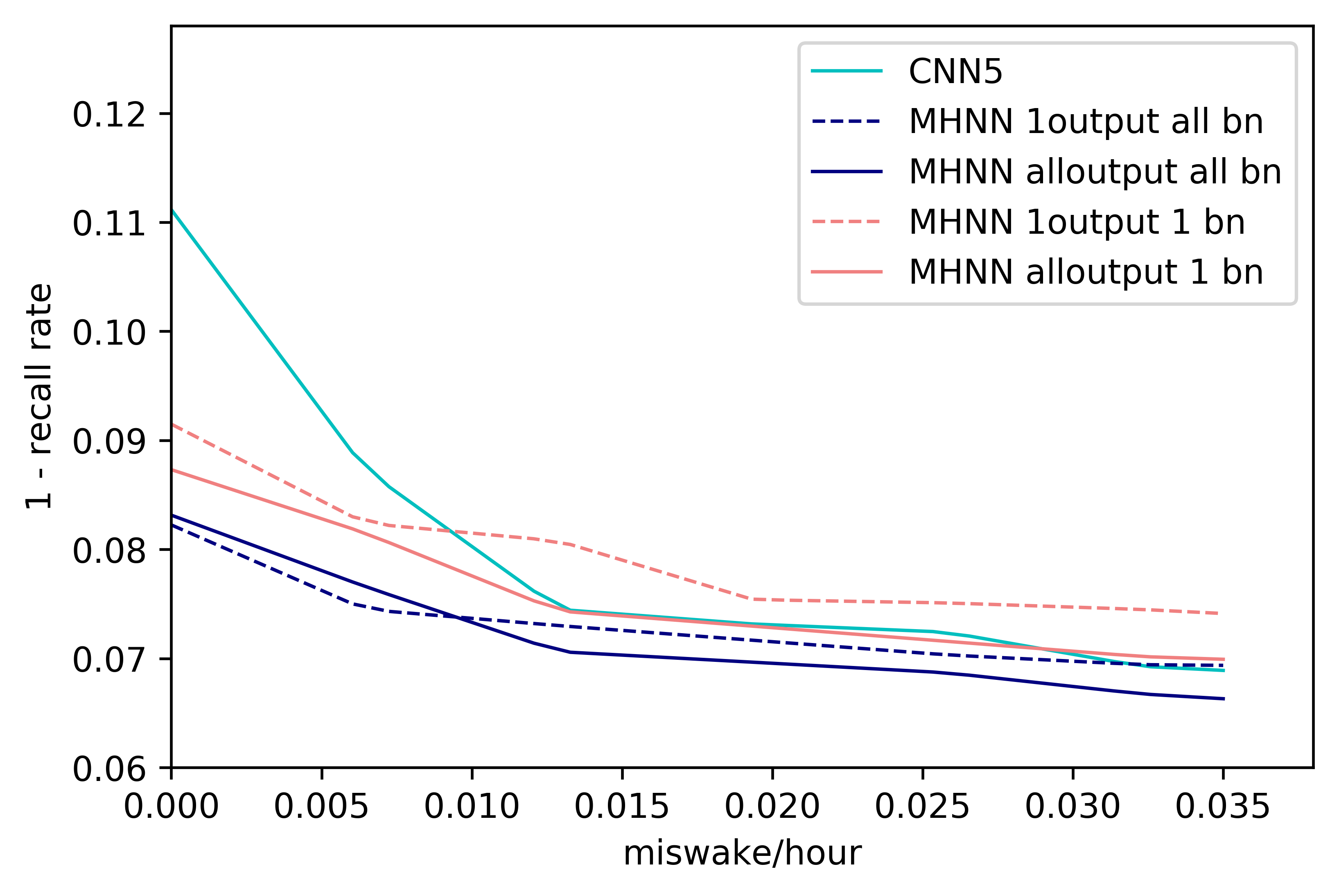}}
 % \vspace{2.0cm}
%  \centerline{(a) Result 1}\medskip
\end{minipage}

\caption{ROC curve of MHNN}%
\label{fig:res}

\end{figure}

According to the ROC, most MHNN performs better than CNN5 which is the best CNN beyond our experiment. 

\section{Conclusion}
\label{sec:majhead}

We have implemented Hierarchical Neural Network into KWS problem. Its performance is better than original DNN and CNN architecture. The model size and computation complexity are low enough to deal with real-time problems. According to past work, CNN baseline performs better than DNN baseline, so we will try to implement three level CNN Hierarchical Neural Network in the future.

\vfill\pagebreak

\section{REFERENCES}
\label{sec:refs}

\noindent [1] Samuel Thomas, Sriram Ganapathy and Hynek Hermansky, ``Multilingual MLP Features For Low-Resource LVCSR System", Proc, of \emph{IEEE ECASSP}, 2012.

\noindent [2] Christian Plahl, Ralf Schl\"{u}ter, Hermann Ney, ``Hierarchical Bettle Neck Feature For LVCSR", in \emph{Interspeech}, 2010.

\noindent [3] Christian Plahl, Ralf Schl\"{u}ter, Hermann Ney, ``Cross-lingual Probability of Chinese and English Neural Network Features For French and German LVCSR", in \emph{IEEE ASRU}, 2011.

\noindent [4] Fabio Valente, Mathew Magimai.-Doss, Christian Plahl, Suman Ravuri, ``Hierarchical Processing of the Modulation Spectrum for GALE Mandarin LVSCR System", in \emph{Interspeech}, 2009.

\noindent [5] David RH Miller, Michael Kleber, Chia-Lin Kai, Owen Kimball, Thomas Colthurst, Stephen A Lowe, Richard M Schwarts and Herbert Gish, "Rapid and Accurate Spoken Term Detection", in \emph{Eighth Annual Conference of the International Speech Communication Association} 2007.

\noindent [6] Siddika Parlak and Murat Saraclar, ``Spoken Term Detection For Turkish Broadcast News", in \emph{Acoustics, Speech and Singal Processing}, 2008, pp. 5244-5247

\noindent [7] Richard C Rose and Douglas B Paul, ``A Hidden Markov Model Based Keyword Recognition System", in \emph{Acoustics, Speech and Signal Proccessing} 1990.

\noindent [8] Jay G Wilpon, Lawrence R Rabiner, C-H Lee, and ER Goldman, "Automatic Recognition of Keywords in Unconstrained Speech Using Hidden Markov Model", \emph{IEEE Transactions on Acoustics, Speech and Signal Processing}, vol. 38, no.11, pp. 1870-1878, 1990.

\noindent [9] Geoffrey Hinton, Li Deng, Dong Yu, George E Dahl, Abdel-rahman Mohamed, Navdeep Jaitly, Andrew Senior, Vincent Vanhoucke, Patrick Nguyen, Tara N Sainath, et al. ``Deep Neural Networks for Acoustic Modeling in Speech Recognition: The Shared Views of Four Research Groups", \emph{IEEE Signal Processing Magazine}, vol.29, no.6, pp.82-97, 2012.

\noindent [10] Sankaran Panchapagesan, Ming Sun, Aparna Khare, Spyros Matsoukas, Arindam Mandal, Bj\"{o}rn Hoffmeister and Shiv Vitaladevuni, ``Multi-task Learning and Weighted Cross-entropy for DNN-based Keyword Spotting", \emph{Interspeech}, 2016.

\noindent [11] Ming Sun, David Snyder, Yixin Gao, Varun Nagaraja, Mike Rodehorst, Sankaran Panchapagesan, Nikko Strom, Spyros Matsoukas and Shiv Vitaladevuni, ``Compressed Time Delay Neural Network for Small Footprint Keyword Spotting", \emph{Interspeech}, 2017.

\noindent [12] Guoguo Chen, Sanjeev Khudanpur, Daniel Povey, Jan Trmal, David Yarowsky and Oguz Yilmaz, ``Quantifying the Value of Pronunciation Lexicons for Keyword Search In Low Resource Languages", \emph{IEEE international Conference on Acoustics}, 2013.

\noindent [13] Daniel Provey et al. ``The Kaldi Speech Recognition Toolkit", \emph{IEEE AERU, IEEE Signal Processing Society}, 2011. 

%List and number all bibliographical references at the end of the
%paper. The references can be numbered in alphabetic order or in
%order of appearance in the document. When referring to them in
%the text, type the corresponding reference number in square
%brackets as shown at the end of this sentence \cite{C2}. An
%additional final page (the fifth page, in most cases) is
%allowed, but must contain only references to the prior
%literature.

\bibliographystyle{IEEEbib}
\bibliography{strings,refs}

\end{document}